\documentclass[journal]{IEEEtran}

\usepackage[utf8]{inputenc}
\usepackage{graphicx}
\usepackage{url}
\usepackage{subcaption}
\usepackage{multirow}
\usepackage{array}
\usepackage{color}
\usepackage{amsmath}
\usepackage{amssymb}

\usepackage{lipsum}                     
\usepackage{xargs}                      
\usepackage[pdftex,dvipsnames]{xcolor}  
\usepackage[colorinlistoftodos,prependcaption,textsize=tiny]{todonotes}
\newcommandx{\unsure}[2][1=]{\todo[linecolor=red,backgroundcolor=red!25,bordercolor=red,#1]{#2}}
\newcommandx{\change}[2][1=]{\todo[linecolor=blue,backgroundcolor=blue!25,bordercolor=blue,#1]{#2}}
\newcommandx{\info}[2][1=]{\todo[linecolor=OliveGreen,backgroundcolor=OliveGreen!25,bordercolor=OliveGreen,#1]{#2}}
\newcommandx{\improvement}[2][1=]{\todo[linecolor=Plum,backgroundcolor=Plum!25,bordercolor=Plum,#1]{#2}}
\newcommandx{\thiswillnotshow}[2][1=]{\todo[disable,#1]{#2}}

\ifCLASSINFOpdf
\else
\fi
\usepackage{algorithmic}
\begin{document}
%
\title{Grab, Pay and Eat: Semantic Food Detection\\
for Smart Restaurants}
%
%
%

\author{Eduardo Aguilar, Beatriz Remeseiro, Marc Bola\~nos, and Petia Radeva, \textit{Fellow, IAPR}
\thanks{Manuscript received January 1, 2020; revised January 1, 2020. This work was partially funded by TIN2015-66951-C2-1-R, SGR 1219, and CERCA Programme / Generalitat de
Catalunya. E. Aguilar acknowledges the financial support of CONICYT Becas Chile, B. Remeseiro acknowledges the support of the Spanish Ministry of Economy and Competitiveness under \textit{Juan de la Cierva} Program (ref. FJCI-2014-21194), and M. Bola\~nos acknowledges the support of an FPU fellowship (ref. FPU15/01347). P. Radeva is partially supported by ICREA Academia 2014. The funders had no role in the study design, data collection, analysis, and preparation of the manuscript.}
\thanks{E. Aguilar is with the Departamento de Ingeniería de Sistemas y Computación, Universidad Católica del Norte, Avenida Angamos 0610, Antofagasta, Chile. B. Remeseiro, M. Bola\~nos and P. Radeva are with the Departament de Matem\`atiques i Inform\`atica, Universitat de Barcelona, Gran Via de les Corts Catalanes 585, 08007 Barcelona, and Computer Vision Center, Cerdanyola (Barcelona), Spain. (e-mail: eaguilar02@ucn.cl, bremeseiro@uniovi.es, marc.bolanos@ub.edu, petia.ivanova@ub.edu.)}}

\maketitle

\begin{abstract}
The increase in awareness of people towards their nutritional habits has drawn considerable attention to the field of automatic food analysis. Focusing on self-service restaurants environment, automatic food analysis is not only useful for extracting nutritional information from foods selected by customers, it is also of high interest to speed up the service solving the bottleneck produced at the cashiers in times of high demand. In this paper, we address the problem of automatic food tray analysis in canteens and restaurants environment, which consists in predicting multiple foods placed on a tray image. We propose a new approach for food analysis based on  convolutional neural networks, we name Semantic Food Detection, which integrates in the same framework food localization, recognition and segmentation. We demonstrate that our method improves the state of the art food detection by a considerable margin on the public dataset UNIMIB2016 achieving about 90\% in terms of F-measure, and thus provides a significant technological advance towards the automatic billing in restaurant environments.
\end{abstract}

\begin{IEEEkeywords}
food tray analysis, food recognition, semantic segmentation, convolutional neural networks
\end{IEEEkeywords}

%
\IEEEpeerreviewmaketitle

\section{Introduction}


%
%
%
%


\IEEEPARstart{H}{aving} a poor routine of physical exercises and poor nutritional habits are two of the main possible causes of people's health-related issues like obesity or diabetes, among others. For these reasons, nowadays people are more concerned about these aspects of their daily life. Therefore, the need for applications that allow to keep track of both physical activities and nutrition habits are rapidly increasing, a field in which the automatic analysis of food images plays an important role. Focusing on self-service restaurants,  food recognition algorithms could enable both monitoring of food consumption and the automatic billing of the meal grabbed by the customer. The latter is quite relevant because remove the need for a manual selection of the chosen dishes, allowing to speed-up the service offered by these restaurants. 

\begin{figure}[htb]
\centering
\includegraphics[width=0.49\columnwidth]{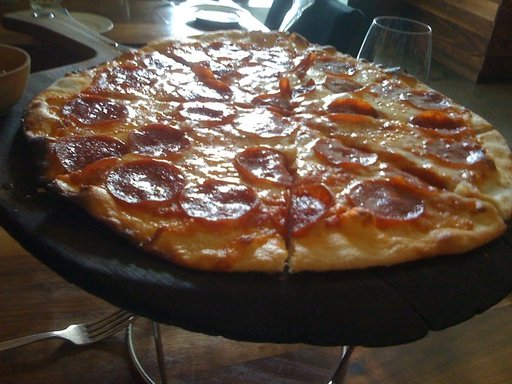} \hfill
\includegraphics[width=0.49\columnwidth]{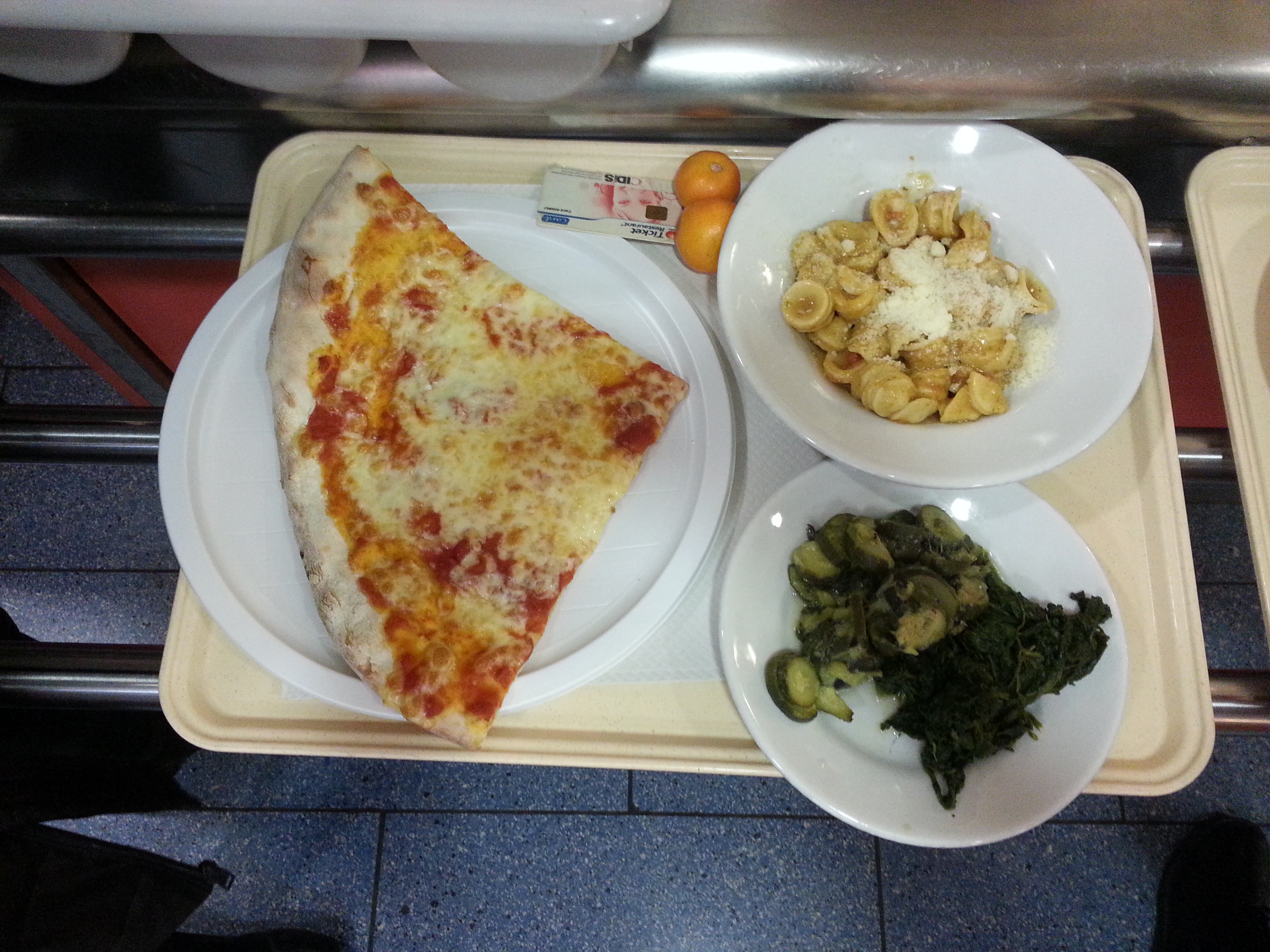}
\caption{Examples of images used in traditional approaches to food analysis (left) and food tray analysis (right).}
\label{fig:problem}
\end{figure}

From the computer vision side, several approaches have been proposed in order to tackle the problem, most of them using Convolutional Neural Networks (CNNs) \cite{kagaya2015,ragusa2016,aguilar2017,singla2016}. Several of the published work consider the development of methods for food recognition, that is, being able to recognize the dish depicted in a picture in which a single plate is shown. An important consideration to take into account when modeling visual food-related information is its fine-grained nature, meaning that specially in the problem of food analysis the intra-class variability and inter-class similarity are hardly making difficult the problem of  obtaining robust food recognition methods.

Several works in the literature have proposed methods usable for applications related to food intake self-monitoring \cite{aizawa2013food,tanno2016deepfoodcam,waltner17b}, in which the user should take pictures of each meal and the system would consequently track any nutritional information associated. Other approaches related to the problem of food intake include a method to assess meal images by food portion estimation using two images acquired by mobile devices \cite{dehais2017two}, a model to learn food ingredients from recipes using state-of-art CNNs as multi-label predictors \cite{bolanos2017food}, and a multimodal multitask deep belief network to learn both visual information and image-ingredient representation \cite{min2017being}.

\begin{figure}[htb]
\centering
\includegraphics[width=.8\columnwidth]{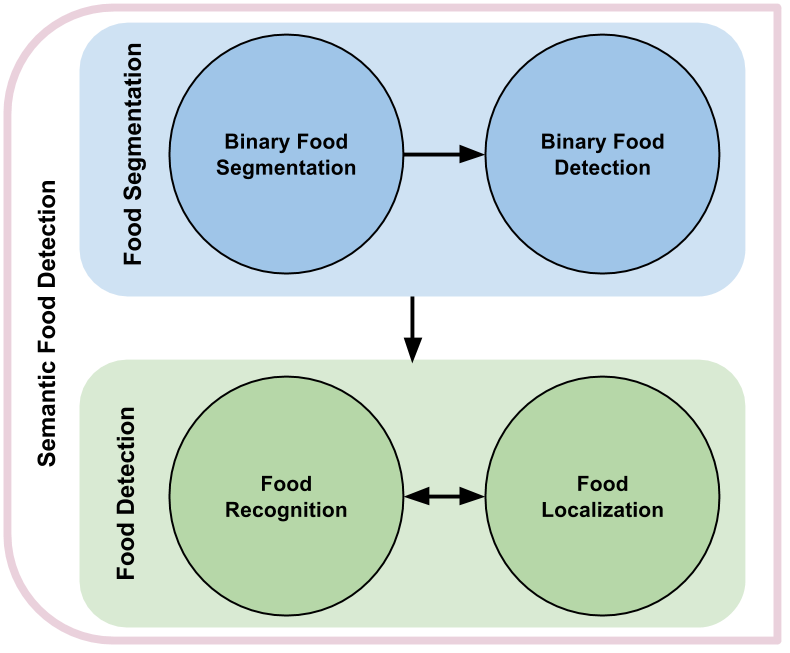}
\caption{Different tasks involved in our Semantic Food Detection framework.}
\label{fig:basic_pipe}
\end{figure}

Instead of applying a personalized tracking, there are several contexts where social monitoring or recognition is required. A clear example might be food tray detection in public spaces \cite{ciocca2015food,ciocca2017food}, where the sample consists of a tray picture that includes all the food that a user is about to consume (see Fig. \ref{fig:problem}) and the model is intended to process all pictures from any possible users taking food at the same restaurant. The development of a system able to apply food tray detection in a controlled, but social and public environment could enable several applications. The most straightforward context of applicability would be automatic billing in self-service restaurants, where the system could solve the need for a person selecting what the customer grabbed before paying. A different application could consider the design of smart trays \cite{raimato2017design}, which could provide food recommendations depending on what the customer is selecting. The provided recommendations could be based on calorie counting, healthy food, specific nutritional composition, etc. In addition, if we also consider a system able to log the food consumed by every individual along time, it could provide health-related recommendations in a long-term way.

There are several aspects that make the food tray analysis a challenging problem \cite{ciocca2017food} like: 1) multiple foods placed on the dishes and placemats, 2) different foods served in the same dish, 3) visual distortions as well as illumination changes due to shadows, and 4) objects placed on a tray that do not correspond to a type of food. On the other hand, unlike traditional approaches to food analysis, difficulties due to intraclass variability have less influence on the problem of food tray detection.

In this work, we propose a novel method that unifies the problems of food detection, localization, recognition and segmentation into a new framework that we call Semantic Food Detection. As Fig. \ref{fig:basic_pipe} shows, to achieve this target, we integrate the information extracted by two main approaches: a) food segmentation and b) object detection trained for food detection, by taking advantage of the benefits provided by both algorithms in a CNN framework. The first one, allows us to determine where the food is in terms of pixel and bounding boxes. The second one allows us to locate and recognize the foods present in the images.
The Semantic Food Detection framework combines the information that provides both algorithms in order to  prevent false food detections and thus provide a better performance.

The contributions of our paper are as follows: 1) a novel framework that integrates the problems of food detection, localization, recognition and segmentation; and 2) a novel approach to address the problem of food tray analysis, that integrates  a fully-convolutional network for semantic segmentation and a convolutional neural network for object detection. Our method achieves about 90\% in terms of F2-score and is able to outperform the state of the art methods by more than 10\% with respect to recall and more than 20\% with respect to mean average accuracy.

The remainder of this paper is organized as follows: Section \ref{sec:related_work} includes an overview of the related work, Section \ref{sec:methods} presents the proposed Semantic Food Detection approach, Section \ref{sec:results} shows the experimental results and discussion, and Section \ref{sec:conclusion} closes with the conclusions and future research.

\section{Related Work} \label{sec:related_work}

Nowadays, there is a great interest in conducting research oriented to the visual food analysis, emphasizing mainly in its applicability for monitoring the diet of the user based on the intrinsic nutritional information contained in food images. In this field, researchers have focused on several aspects related to automatic food analysis.

The most basic aspect tackled in the literature is the \emph{binary food detection} problem that determines the presence or absence of food in an image. This problem is also called food/non-food classification or food detection \cite{kagaya2014}. The first approximation was proposed by Kitamura et al. \cite{kitamura2009}, who through the combination of a BoF model and an SVM achieve a high accuracy on a tiny dataset of 600 images. An improvement of about 4\% is achieved in terms of overall accuracy using a method based on CNN \cite{kagaya2014}. From this, numerous researchers have proposed models based on CNNs either for feature extraction \cite{ragusa2016,aguilar2017} or for the whole recognition process \cite{kagaya2015,singla2016}. The best results obtained on public datasets with more than 15.000 images \cite{kagaya2015,ragusa2016} have been reported in \cite{aguilar2017} through the combination of CNN GoogLeNet for feature extraction, PCA for dimensionality reduction and SVM for classification. As for its applicability, this problem has commonly been investigated for indexing WEB images \cite{kitamura2009} or as a pre-processing method for an automatic food recognition system \cite{kagaya2015,singla2016}. It has also been used to detect bounding boxes in an image, where food is present \cite{bolanos2016} and to automate the process of image cleaning required when gathering the images of a food dataset \cite{mezgec2017}.

\begin{figure*}[htb]
\centering
\includegraphics[width=\textwidth]{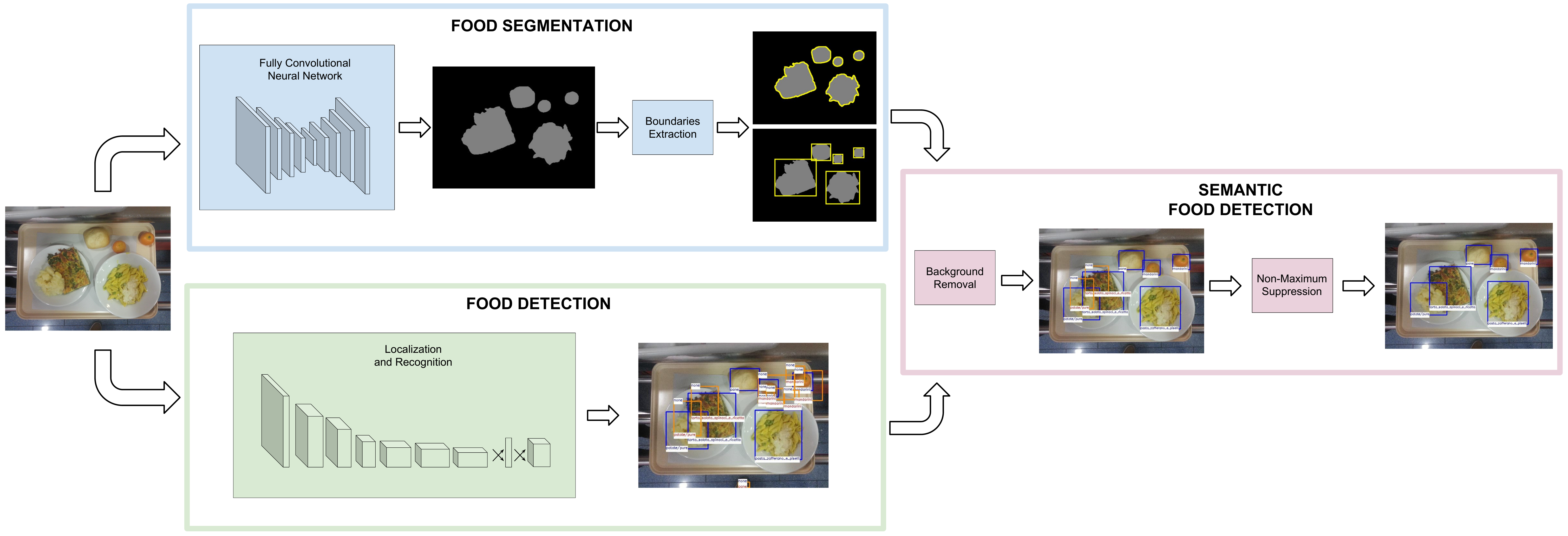}
\caption{Detailed workflow of the proposed Semantic Food Detection method: food segmentation and food detection methods are applied in parallel, before combining them for a final detection on food tray images.}
\label{fig:main_pipe}
\end{figure*}

In food analysis, once images containing food are identified, \emph{food recognition} is usually the next step to apply. Again, CNN-based models have been able to progressively improve the results of food recognition models reaching an accuracy of about 90\% in datasets with around 100 different kinds of food \cite{martinel2016}. In general, the best proposals are based on the winning models of the ILSVRC challenge \cite{russakovsky2015}, and a fine-tuning process is usually applied either making some architectural model changes (e.g. addition or removal of layers) \cite{yanai2015,liu2016} or not \cite{hassannejad2016}. Several datasets have been proposed for tackling this problem: a) datasets including fine-grained classes (e.g. apple pie, pork chop, pizza, tiramisu), being the most popular ones UECFOOD-100 \cite{matsuda2012}, UECFOOD-256\cite{kawano2014} and Food-101\cite{bossard2014}, and b)  datasets based on high-level categories (e.g. dessert, meat, vegetables, soup), like Food-11 \cite{singla2016}. The best result when using fine-grained classes was achieved by the WISeR model \cite{martinel2016}, which combines the food traits and the vertical structure of some food, extracted by the standard squared convolutional kernel and the proposed slice convolutional kernel, respectively. Regarding the high-level categories, the best results were obtained by \cite{aguilar2017_2} through a novel proposal that fuses several CNN models, achieving a 10\% improvement in terms of accuracy with respect to the baseline method through the combination of the outputs generated by two CNN models (GoogLeNet and ResNet50).

Most of the approaches focused on food recognition only exploit the visual content, but they ignore the context. However, geolocation and other information have also been explored in the literature for restaurant-oriented food recognition: on-line restaurant information is used in \cite{bettadapura2015leveraging}, similarly to \cite{beijbom2015menu} in which nutritional information is also retrieved; whilst the menu, the location and user images of dished are used in \cite{xu2015geolocalized}. On the other hand, Herranz et al. \cite{herranz2017modeling} go a step further since their target is not only to improve both classification performance and efficiency, but also to better model contextual data and its relation with the other elements.

To date, most food recognition algorithms and datasets focus on classifying images that include only one dish \cite{martinel2016, liu2016, hassannejad2016}. However, in some cases, there may be more than one dish in the image and, in some cases, the dish can contain several kinds of food. \emph{Food localization} and \emph{food segmentation} are two tasks intended to cope with these problems. The former consists in extracting the regions of the images where the food is located. Up to our knowledge, the only available approach that does not require segmenting the food before extracting the bounding boxes is the one proposed by \cite{bolanos2016}. The task of food segmentation consists in classifying each pixel of the images representing a food. The latest research for food segmentation proposes an automatic weakly supervised methods \cite{shimoda2015,shimoda2016}, which are based on Deep Convolutional Neural Networks and Distinct Class-specific Saliency Maps, respectively.

In this paper, we deal with the identification of different foods placed on a food tray, by integrating the four food analysis problems mentioned above. To the best of our knowledge, only one approach with this purpose has been evidenced in the literature \cite{ciocca2017food}. The authors there introduced an additional food dataset composed by images taken in a canteen environment named UNIMIB2016. In addition, they proposed a pipeline for food recognition that performs classification based on the candidate regions obtained by combining two separate images segmentation processes, through saturation and color texture (JSEG). The best result was achieved by combining global and local (path-based) approaches using an SVM as classifier. 
However, the proposed approach performs the segmentation based on generic methods rather than learning the best discriminant features between different foods based on the dataset. Furthermore, it requires several sequential steps to first segment and then classify, which implies a high processing time. Instead, our method is able to perform the food segmentation and detection processes in parallel, allowing to speed up the processing time.

\section{Semantic Food Detection} \label{sec:methods}

This work proposes a method for food tray semantic detection that integrates food vs non-food semantic segmentation with food localization and recognition. The pipeline of our approach is given in Fig. \ref{fig:main_pipe} and explained in detail in the following subsections.

\subsection{Food Segmentation} \label{sec:methods:segmentation}

Food segmentation deals here with the problem of separating the food and food-related items, from the tray and other background elements, thus obtaining a binary image. For this purpose, we apply semantic segmentation techniques that work in a supervised learning framework, unlike the most segmentation methods that focus on image properties (e.g. color or texture). Notice that semantic segmentation could be used to directly segment the input image into the different food categories. However, the most recent methods in this field provide great results with datasets that contain a relatively low number of classes, such as CamVid \cite{brostow2008segmentation} with 11 semantics classes or Gatech \cite{hussain2013geometric} with 8. The number of categories used in food analysis is much higher, thus increasing the difficulty of the task and providing not so satisfactory results \cite{shelhamer2017fully}.

Fully convolutional networks (FCNs) \cite{shelhamer2017fully} are the state-of-the-art in the field of semantic segmentation. FCNs are composed of convolutional layers only, which means that they do not have any fully-connected layer. FCNs take input images of arbitrary size and produce outputs of the same size by means of an efficient inference and learning process.

Several FCN-models can be found in the literature applied to semantic segmentation. One of the most notable examples with interesting image segmentation results is the Tiramisu model \cite{jegou2017one}. As any FCN, Tiramisu is composed of a down-sampling path and an up-sampling path, which, in this case, are connected by skip connections. Its architecture is additionally composed of dense blocks, based on the idea of densely connected convolutional networks (DenseNets) \cite{huang2017densely}, each one of them containing a set of concatenated layers.

Given the binary image predicted by the FCN model, the next step aims at tracing the exterior boundaries of the food regions, avoiding the holes inside them. In this manner, small holes that may appear inside regions are discarded in this stage and thus the regions are homogenized. For this task, we use the Moore-Neighbor tracing algorithm modified by Jacob's stopping criteria \cite{gonzalez2004digital}.

\begin{figure}[htb]
\centering
\includegraphics[width=0.32\columnwidth]{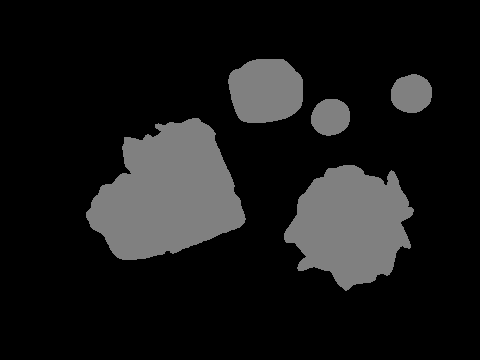} \hfill
\includegraphics[width=0.32\columnwidth]{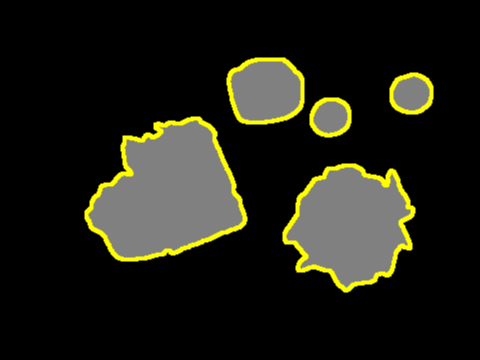} \hfill
\includegraphics[width=0.32\columnwidth]{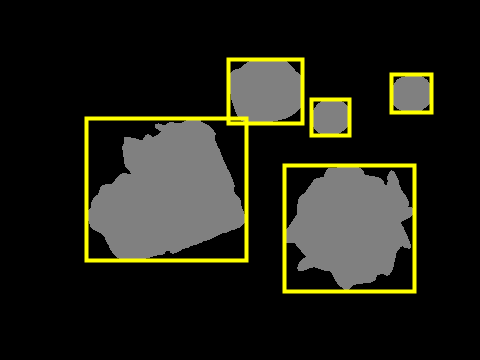} \\ \vspace*{0.2cm}
\caption{Food segmentation outputs, from left to right: binary image predicted by the FCN model, boundaries of the regions, and food bounding boxes.}
\label{fig:segmentation}
\end{figure}

Once the boundaries are traced, the bounding boxes that contain the regions are determined, thus obtaining a binary food detection. As small regions may also appear in the predicted images, and they usually correspond to false positives, this step also includes their elimination by considering a threshold criterion.
Figure \ref{fig:segmentation} illustrates an example of the outputs obtained in the food segmentation procedure, including the binary image provided by the FCN model, the boundaries extracted and the bounding boxes generated.

\subsection{Food Detection}
In this work, following the definition of the Object Detection problem 
\cite{russakovsky2015}, we consider as Food Detection the localization and recognition of food. 
For this purpose, we propose re-training an object detection algorithm for applying food detection instead. YOLOv2 \cite{Redmon2016you, Redmon2016} is currently one of the best object detection approaches in the state of the art. It allows to predict the bounding boxes and class probability of any object with a single convolutional neural network in real time. 
As for the model, the authors propose a new FCN called Darket-19, composed by 19 convolutional layers and 5 max pooling layers to tackle the recognition task. This network is modified for object detection by: 1) removing the last convolutional layer and adding four convolutional layers for producing 13x13 feature maps, and 2) providing a region selection that enables to predict $B$ bounding boxes at each cell on the output feature maps. 
The network predicts five coordinates for each bounding box, among them is the confidence score $t_o$, which represents both the confidence that the box contains an object and the accuracy at which the object is believed to be predicted; and $c=1, \dots, C$ conditional class probabilities, $Pr(Class_c|Object)$. Predictions are obtained from the last convolutional layer having a size equal to $1\times1$ and $F$ filters, where the number of filters is calculated as: $F =(B \times (5 + C))$. 
From this, it is possible to determine the class-specific confidence score, $CS_c$ for each bounding box as follows:  
\begin{equation}
\begin{aligned}
CS_c= & Pr(Class_c|Object) * \sigma(t_o)
\end{aligned}
\end{equation}
where $\sigma(.)$ stands for a logistic activation to constrain the predictions to fall in the range between 0 and 1.

\begin{figure*}[htb]
\centering
\begin{subfigure}{0.49\columnwidth}
\includegraphics[width=\columnwidth]{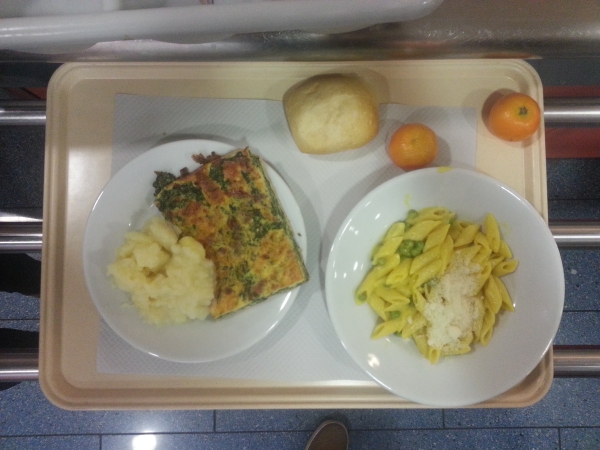}
\caption{} 
\end{subfigure} \hfill
\begin{subfigure}{0.49\columnwidth}
\includegraphics[width=\columnwidth]{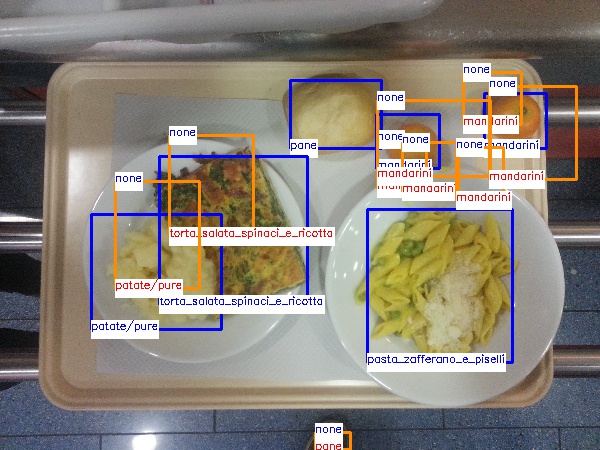}
\caption{} 
\end{subfigure} \hfill
\begin{subfigure}{0.49\columnwidth}
\includegraphics[width=\columnwidth]{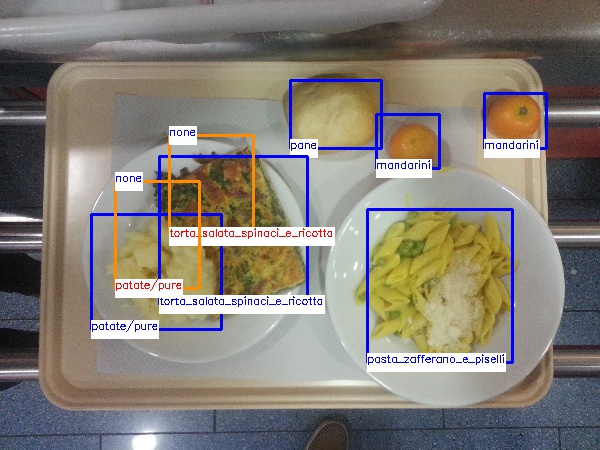}
\caption{} 
\end{subfigure} \hfill
\begin{subfigure}{0.49\columnwidth}
\includegraphics[width=\columnwidth]{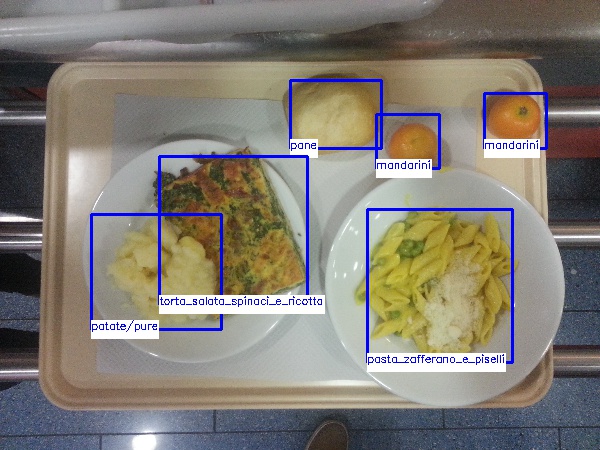}
\caption{} 
\end{subfigure} 
\caption{Semantic Food Detection outputs: a) original image, b) the results of the food detection with YOLOv2, c) the results of applying Background Removal procedure to the bounding boxes detected, and d) the results after applying Non-Maximum Suppression to the remaining bounding boxes. Note that blue bounding boxes stand for correct food detections, and orange for false food detections.}
\label{fig:detection:input-yolo}
\end{figure*}

Figure \ref{fig:detection:input-yolo} (b) illustrates the bounding boxes and classes extracted with YOLOv2 using the threshold $\frac{1}{65}$. We can observe that all dishes have been detected, but we got eight false detections. Most of the false detections were located around the plates (duplicate detections) and only one was completely in the background.


\subsection{Semantic Food Detection}


In object detection, one of the most common errors are false positives, which can be classified based on the type of error: localization error, confusion with similar objects, confusion with dissimilar objects, and confusion with background \cite{hoiem2012diagnosing}. Our Semantic Food Detection proposal focuses on reducing two of the most common errors of object detectors \cite{Redmon2016you}: localization errors, specifically those corresponding to duplicate detections; and errors produced by the confusion with the background. For this purpose, we propose the following procedure that integrates the detection and segmentation algorithms:

\subsubsection{Background Removal}

The first step involves the application of both boundaries extracted (contour and bounding box) from the Food Segmentation procedure in order to remove the background detections. Let $Y=\{b^Y_1, \dots, b^Y_N\}$ be the set of bounding boxes obtained with the detection method,
$S_1=\{b^S_1, \dots, b^S_L\}$ and $S_2=\{c^S_1, \dots, c^S_L\}$ the set of bounding boxes and contours extracted by the Food Segmentation method, respectively. Considering that each element belonging to the sets named above can be considered as a set of points $(x,y)$ that defines a polygon, we calculate the probability of a bounding box, $b^Y_i$ to belong to the background $Bkg$ as follows:


\begin{equation*}
\begin{aligned}
& Pr(Bkg|b^Y_i) = \min(CS_c(\overline{b^Y_i}),  \max(Pr(\overline{S_1|b^S_i}) Pr(\overline{S_2|b^Y_i}))
\end{aligned}
\end{equation*}
where $CS_c(\overline{b^Y_i})$ is 
the complement of the confidence score ($1-CS_c(b_i^Y)$) for the i-th detection, $Pr(\overline{S_1|b^Y_i})$ denotes the probability that $b^Y_i$ is a false detection based on the extracted boxes ($S_1$):
\begin{equation*}
\begin{aligned}
& Pr(\overline{S_1|b^Y_i}) = \min_{j=1,\dots,L }\frac{|b^Y_i \cap b^S_j|}{|b^Y_i|}
\end{aligned}
\end{equation*}
where $|.|$ stands for the cardinality of a set of pixels corresponding to an image region, and $Pr(\overline{S_2|b^Y_i})$ the probability that $b^Y_i$ does not intersect with any contour in $S_2$:

\begin{equation*}
\begin{aligned}
& Pr(\overline{S_2|b^Y_i}) = \min_{j=1,\dots,L} Ind(b^Y_i \cap c^S_j = \emptyset)
\end{aligned}
\end{equation*}
where $Ind(*)$ is an indicator function with value 1 if the condition is true, and 0 otherwise.


Bounding boxes with a probability higher than 50\% to be background ($Pr(Bkg|b_i^Y)> T, T=0.5$) are considered to be false detections, and are therefore removed. 

Figure \ref{fig:detection:input-yolo} (c) illustrates an example of the results obtained after applying this procedure. As it can be observed, false detections around the objects of the class {\em mandarine} have been removed.


\subsubsection{Non-Maximum Suppression}
The second step involves the application of a greedy procedure to eliminate duplicate detections by non-maximum suppression  \cite{Felzenszwalb2009}. Once the Background Removal is applied, the remaining detections $Y' \subseteq Y$ are sorted in descending order by the confidence score $CS_c(b^Y_j)$ and grouped into $C$ sets $Y^{1},\dots, Y^{C} \subset Y'$, where $C$ is the number of classes. 
Then, for each $Y^{c}, c=1,...C$, we greedily select the highest scoring bounding boxes while removing detections that are lower in the ranking and their maximum intersection with respect to the i-th previously selected bounding boxes is more than 50\%. 

Figure \ref{fig:detection:input-yolo} (d) shows that non-maximum suppression procedure is able to eliminate the remaining false detections while keeping the foods well localized and classified.


\section{Experimental Results} \label{sec:results}

In this section, we first describe the dataset used to evaluate the proposed approach, which is composed of images taken in self-service restaurants. Then, we describe the evaluation measures used and present the results obtained with the different methods and model configurations.

\subsection{Dataset}

UNIMIB2016 \cite{ciocca2017food} is a food dataset that has been collected in a self-service canteen. Each image includes a tray with some food placed both on plates and placemats. The acquisition process was performed on a semi-controlled environment using a Samsung Galaxy S3 smartphone. As a result, images acquired have a resolution of $3264 \times 2448$ in RGB, and present visual distortions and variable illuminations, making them challenging for any task of automatic food analysis.

The dataset is composed of 1.027 images that include a total of 73 food categories. Among them, only 1.010 images and 65 categories were used for experimentation, as suggested in \cite{ciocca2017food} due to the low number of samples of the categories not considered. For experimental purposes, the dataset has been split in training and test sets: the former contains 650 images ($\approx 64\%$), whilst the latter contains 360 ($\approx 36\%$).

\begin{figure}[htb!]
\centering
\includegraphics[width=0.44\columnwidth]{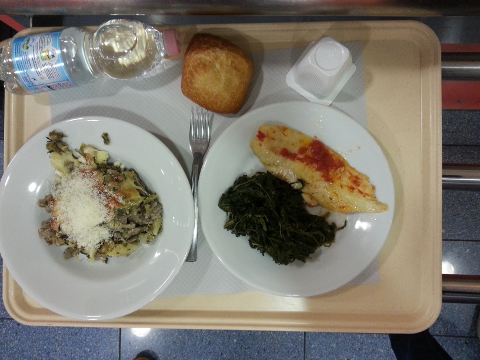} \hspace*{0.2cm}
\includegraphics[width=0.44\columnwidth]{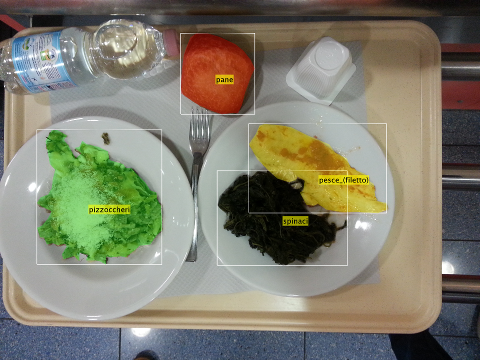}
\caption{A representative sample of the UNIMIB dataset \cite{ciocca2017food}: original image (left) and food annotations (right).}
\label{fig:unimib}
\end{figure}

The annotations included in the dataset contain, for each food item: the polygon defining its boundaries, the bounding box and the food label. Figure \ref{fig:unimib} illustrates an image of the UNIMIB2016 dataset with its corresponding annotations.

\subsection{Food Segmentation}

\textbf{Metrics.}
In order to evaluate the different food segmentation approaches, several performance measures have been used. First, two pixel-wise metrics commonly used in semantic segmentation problems have been considered \cite{shelhamer2017fully}:

\begin{itemize}
\item \emph{Global pixel accuracy}. The pixel-wise accuracy computed over all the pixels of the dataset.
\item \emph{Intersection over Union (IoU)}. Also known as Jaccard index, it is defined as:
\begin{equation}
IoU(c) = \frac{\sum_i t_i==c \wedge p_i==c}{\sum_i t_i==c \vee p_i==c}
\end{equation}
where $c$ is a class, $i$ represents all the pixels of the dataset, $t_i$ are the target labels, and $p_i$ are the predicted labels. Note that this metric is calculated for each single class $c$, and then the mean across the classes is computed.
\end{itemize}

With the main aim of making a fair comparison with the results presented in \cite{ciocca2017food}, three region-based metrics have been also considered \cite{arbelaez2011contour}:

\begin{itemize}
\item \emph{Covering}. The covering of the ground truth ($GT$) by the segmented ($S$) images measures the level of overlapping between each pair of regions ($R$ and $R'$):

\begin{equation}
C(S \rightarrow GT) = \frac{1}{N} \sum_{R \in GT}^{} |R| \cdot \underset{R' \in S}{max} \frac{|R \cap R'|}{|R \cup R'|}
\end{equation}
where $N$ is the number of pixels of the image.

\item \emph{Rank index}. It compares the compatibility of assignments between pairs of elements in the ground truth ($GT$) and the segmented ($S$) images:
\begin{equation}
\begin{aligned}
& RI(S,GT) = \\
& \frac{1}{\binom{N}{2}} \sum_{i < j}  \left [\mathbb{I}(t_i==t_j \wedge p_i==p_j) +
\mathbb{I}(t_i \neq t_j \wedge p_i \neq p_j)  \right ]
\end{aligned}
\end{equation}
where $\binom{N}{2}$ is the number of possible unique pairs among the $N$ pixels of each image, and $\mathbb{I}$ is the identity function.

\item \emph{Variation of information}. It measures the distance between the ground truth ($GT$) and the segmented ($S$) images in terms of their average conditional entropy:
\begin{equation}
VI(S,GT) = H(S) + H(GT) - 2 \cdot MI(S,GT)
\end{equation}
where $H$ and $MI$ are, respectively, the entropy and the mutual information. In this case, the lower the better.
\end{itemize}
Notice that these three metrics are calculated for each single image, and then the mean across images is computed.

\textbf{Experimental setup.}
Regarding the methods used for semantic segmentation, we trained three networks based on the Tiramisu model \cite{jegou2017one}: 

\begin{itemize}
\item \textit{Tiramisu56}: 56 layers, with 4 layers per dense block and a growth rate of 12.
\item \textit{Tiramisu67}: 67 layers, with 5 layers per dense block and a growth rate of 16.
\item\textit{ Tiramisu103}: 103 layers, with a variable number of layers per dense block (from 12 to 4 in the downsampling path, and from 4 to 12 in the upsampling) and a growth rate of 16.
\end{itemize}

Additionally, the \textit{Classic Upsampling}, which uses standard convolutions in the upsampling path instead of dense blocks \cite{ronneberger2015u}, has been also considered for comparative purposes. 

All the models were initialized with HeUniform \cite{he2015delving} and trained with RMSprop \cite{tieleman2012rmsprop}, with an initial learning rate of $1e-3$ and an exponential decay of 0.995 per epoch. For the pre-training, we used crops of $224\times224$ and batch size 3. Then, the models were fine-tuned with full size images and batch size 1, using a learning rate of $1e-4$. The outputs were monitored using the global accuracy and the IoU, with a patience of 100 during pre-training and 50 during fine-tuning.

Table \ref{tab:ss:comparison} includes the results achieved with the four networks used for semantic segmentation, as well as with the two segmentation methods reported in \cite{ciocca2017food}: the JSEG algorithm \cite{deng2001unsupervised}, and the segmentation pipeline proposed by Ciocca et al. \cite{ciocca2017food}. With respect to the pixel-wise measures, all the networks produced competitive results (over 0.96). The Tiramisu models outperformed the Classic Upsampling, thanks to the dense blocks and despite a lower number of parameters. In general terms, the Tiramisu model benefits from having more parameters and depth. However, in this binary problem the Tiramisu103 produced overfitting whilst the Tiramisu67 achieved the best results, with a good trade-off between depth and performance. Regarding the region-based measures, all the FCNs provided better results than the two approaches reported in \cite{ciocca2017food}, which demonstrated the adequacy of the proposed methods for the problem at hand.

\begin{table*}[!t]
	\renewcommand{\arraystretch}{1.3}
	\caption{Results obtained by our Food Segmentation approach in test set.} \label{tab:ss:comparison}
    \centering
	\begin{tabular}{lccc@{\extracolsep{10pt}}ccc} \hline
    & & \multicolumn{2}{c}{Pixel-wise} & \multicolumn{3}{c}{Region-based} \\ \cline{3-4} \cline{5-7} 
    & No. parameters & Global accuracy & Mean IoU & Covering & Rank index & Variation of info.\\ \hline \hline
	JSEG \cite{deng2001unsupervised} & - & - & - & 0.385 & 0.389 & 3.106 \\
    Ciocca et al. \cite{ciocca2017food} & - & - & - & 0.916 & 0.931 & 0.429\\ \hline
   	Classic Upsampling & 12.7M & 0.991 & 0.962 & 0.984 & 0.982 & 0.125 \\
    Tiramisu56 & 1.4M & 0.992 & 0.967 & 0.986 & 0.984 & 0.112 \\
    Tiramisu67 & 3.5M & \textbf{0.993} & \textbf{0.971} & \textbf{0.987} & \textbf{0.986} & \textbf{0.105} \\ 
    Tiramisu103 & 9.4M & 0.992 & 0.968 & 0.986 & 0.984 & 0.111 \\
\hline
	\end{tabular}
\end{table*}

\subsection{Semantic Food Detection Performance}

\textbf{Metrics.}
In order to evaluate food recognition and localization, we chose three standard measures commonly used in multi-class object recognition problems:

\begin{itemize}
\item \emph{Recall ($Rec$)}. The proportion of true positives detected.
\item \emph{Precision ($Pre$)}. The proportion of the true positives against all the positive results.
\item \emph{$F_{\beta}$-measure}. A weighted average of precision and recall. We use $\beta=2$ ($F_2$) to place more emphasis on wrong classified or undetected foods. 
\end{itemize}

For comparative purposes, the measures used by Ciocca et al. \cite{ciocca2017food} were also considered: 

\begin{itemize}
\item \emph{Standard Accuracy ($SA$)}. It is equivalent to the recall.
\item \emph{Macro Average Accuracy ($MAA$)}. The proportion of correctly classified foods, but taking into account the class imbalance of the dataset: 
\begin{equation}\label{eq:eq8}
MAA = \frac{1}{C}\sum_{c = 1}^C \frac{TP_{c}}{NF_{c}},
\end{equation}
where $C$ is the number of classes, $TP_{c}$ is the number of correctly classified foods of class $c$, and $NF_{c}$ is the total number of foods of class $c$.

\item \emph{Tray Accuracy ($TA$)}. The percentage of trays for which all the foods contained are correctly recognized: 
\begin{equation}\label{eq:eq9}
TA = \frac{1}{T} \sum_{t = 1}^T Ind(\frac{TP_t}{NF_t}=1),
\end{equation}
where $T$ is the number of food tray images, $TP_t$ is the number of correctly classified foods on the tray $t$, and $NF_t$ is the total number of foods on the tray $t$.
\end{itemize}





\textbf{Experimental setup.}
As for the experimental setup, YOLOv2 was pre-trained on the ILSVRC dataset. Following, we adapted it by changing the output of the model to 65 classes and applied a fine-tuning using UNIMIB2016 images. For training the model, we used the framework Darknet \cite{darknet13}. The models were trained during 4000 iterations with a batch size of 32, and a learning rate of $1e-3$. In addition, we applied a decay of 0.9 to the iterations 3000 and 3500. 

Once YOLOv2 training is completed, the next step is to determine the confidence threshold to be used during localization and recognition of the food. A low confidence threshold implies a greater number of detections, which maximizes the likelihood that all the foods present in the image will be detected. At the same time, it also increases the chances of obtaining false detections. Taking into account that the confidence defined by the detection method considers two factors (the fit of the bounding box to the object and the predicted class), we chose the minimum threshold according to the number of classes. Given that the target dataset has 65 classes, the minimum threshold chosen is $\frac{1}{65}$. With this value, it can be interpreted that the bounding boxes extracted will have a recognition probability greater than a random value when the detected bounding box fits the object perfectly. Following the interpretation given, we chose $\frac{1}{2}$ as maximum threshold, which implies a high probability, at least 50\%, that the localized object is correctly classified.

Table \ref{tab:detection:yolothresh} shows the results obtained in the training set using different confidence thresholds. The tested thresholds range from the minimum and maximum values mentioned above. As we can observe, when the threshold increases, the precision also increases considerably, but the rest of the indicators are affected. When comparing the results obtained between YOLOv2 and the proposed method, for the minimum threshold, it can be observed that a significant improvement in precision is obtained ($\approx$40\%) with only a slight decrease in the other indicators (0.1\%-0.2\%). Another interesting aspect to highlight is when comparing the results using the maximum threshold, in which case the results are practically identical for both methods. This means that, for a threshold of $\frac{1}{2}$, there are almost no false detections that can be reduced with our procedure. For the remaining experiments, the minimum threshold was chosen for two main reasons: 1) it obtains the best results for the $Recall$, $MAA$ and $TA$ indicators; and 2) it allows us to discard the false positives that appear when combining the results with the food segmentation procedure.

\begin{table}
	\renewcommand{\arraystretch}{1.5}
	\caption{Results obtained by YOLOv2 and the proposed approach in training set using different confidence thresholds.} \label{tab:detection:yolothresh}
    \centering
	\begin{tabular}{lccccccc} \hline
     & & 1/65 & 1/32 & 1/16 & 1/8 & 1/4 & 1/2 \\ \hline \hline
    \parbox[t]{2mm}{\multirow{4}{*}{\rotatebox[origin=c]{90}{\textbf{YOLOv2 \cite{Redmon2016}}}}}
     & $Pre$ & \textbf{0.511} & 0.687 & 0.832 & 0.926 & 0.968 & 0.994  \\
     & $Rec$ & 0.999 & 0.998 & 0.997 & 0.995 & 0.988 & 0.966  \\ 
     & $MAA$ & 0.999 & 0.997 & 0.995 & 0.992 & 0.981 & 0.952  \\
     & $TA$ & 0.997 & 0.992 & 0.988 & 0.982 & 0.960 & 0.895  \\ \hline
\parbox[t]{2mm}{\multirow{4}{*}{\rotatebox[origin=c]{90}{\textbf{Proposed}}}}
   & $Pre$ & \textbf{0.918} & 0.952 & 0.973 & 0.984 & 0.991 & 0.996  \\
   & $Rec$ & 0.998 & 0.997 & 0.996 & 0.994 & 0.987 & 0.965  \\ 
   & $MAA$ & 0.999 & 0.999 & 0.996 & 0.994 & 0.981 & 0.951  \\
   & $TA$ & 0.995 & 0.992 & 0.988 & 0.982 & 0.957 & 0.894  \\ \hline
	\end{tabular}
\end{table}

\begin{figure}[htb]
\centering
\begin{subfigure}{0.49\columnwidth}
\includegraphics[width=\columnwidth]{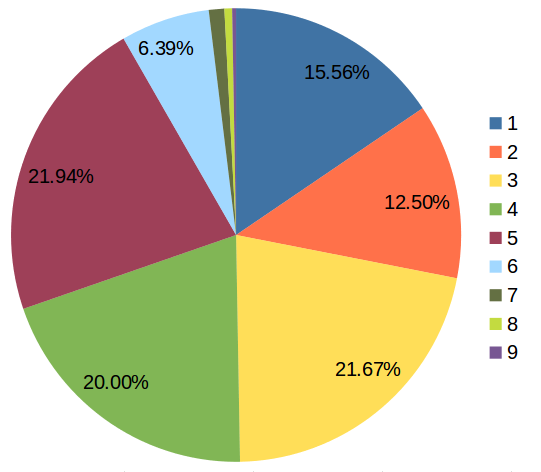}
\caption{} 
\end{subfigure} \hfill
\begin{subfigure}{0.49\columnwidth}
\includegraphics[width=\columnwidth]{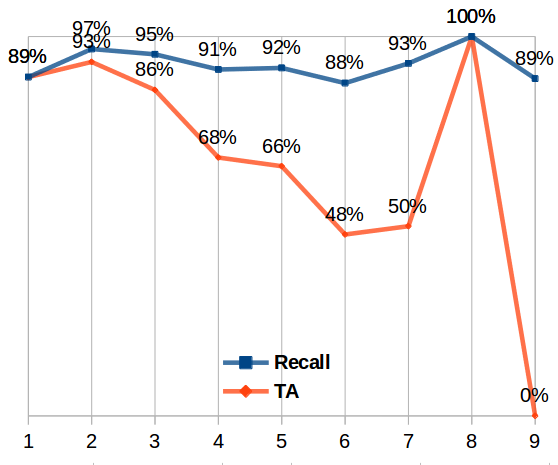} 
\caption{} 
\end{subfigure} \hfill
\caption{a) Distribution of the trays according to the number of foods that are placed in them, and b) Results in terms of \textit{Recall} (blue) and \textit{TA} (orange), for each item of the distribution.}
\label{fig:fbt_results}
\end{figure}

The Semantic Food Detection results on the test set are shown in Table \ref{tab:detection:comparison}. First, it should be highlighted that our proposal outperforms the food recognition, with respect to the state-of-the-art method (Ciocca et al. \cite{ciocca2017food}) in a 12.4\% for \textit{Recall} and 20.9\% for\textit{ MAA}. Regarding \textit{TA}, a decrease of 2.5\% is observed. However, we consider that this measure does not reflect how well the recognition works mainly due to the imbalance in the quantity of food in the trays, which varies between 1 and 9 (see Fig. \ref{fig:fbt_results} (a)), as well as because \textit{TA} measures the amount of food trays in which all positive samples have been correctly predicted, but does not penalize when there are false positives. 

In order to apply a complete comparison, we also replicated the evaluation proposed by \cite{ciocca2017food}, in which the authors considered a perfect segmentation using the ground truth (GT) and applied their detection method (bottom section of Table \ref{tab:detection:comparison}). In our case, there is no significant improvement with respect to the use of the proposed semantic segmentation, because our proposal considers the integration of the extracted information with the segmentation to refine the predictions already obtained by the object detection method. 
In contrast, Ciocca et al. \cite{ciocca2017food} performed the recognition directly on the segmented objects. Comparing to the results obtained in \cite{ciocca2017food}, we can see that their method improves significantly in terms of \textit{Recall} using the GT for segmentation, achieving to match our results. However in terms of \textit{MAA}, despite improving its performance, our results are still about 16\% better. A low \textit{MAA} with a high \textit{Recall} implies that the classifier has a strong bias towards the classes that have a greater amount of instances. Therefore, even if we consider a perfect segmentation to contrast the results, our proposal keeps a better performance in the recognition and, in particular, a lower bias towards the dominant classes.

\begin{table}
	\renewcommand{\arraystretch}{2}
	\caption{Tray Food Analysis Results, from top to bottom: results of YOLOv2 re-trained for food detection, results of the state-of-the-art method and our proposal, and results achieved considering the ground-truth segmentation to perform the recognition. The best results are in boldface.} \label{tab:detection:comparison}
    \centering
	\begin{tabular}{lccccc} \hline
    & $F_2$ & $Pre$ & $Rec$ & $MAA$ & $TA$ \\ \hline
    \hline
    YOLOv2 \cite{Redmon2016} & 0.786 & 0.489 & 0.927 & 0.850 & 0.769  \\ \hline
    Ciocca et al. \cite{ciocca2017food} & - & - & 0.798 & 0.636 & \textbf{0.789}  \\ 
   \textbf{Proposed} & \textbf{0.905} & \textbf{0.843} & \textbf{0.922} & \textbf{0.845}& 0.764 \\ \hline
   Mezgec et al. \cite{mezgec2017} & - & - & 0.864 & - & - \\
   Ciocca et al. \cite{ciocca2017food} & - & - & 0.891 & 0.684 & \textbf{0.871}  \\ 
   \textbf{Proposed} & \textbf{0.911} & \textbf{0.857} & \textbf{0.925} & \textbf{0.849} & 0.772 \\ 
\hline
	\end{tabular}
\end{table} 

\begin{figure*}[htb]
\centering
\includegraphics[width=0.325\textwidth]{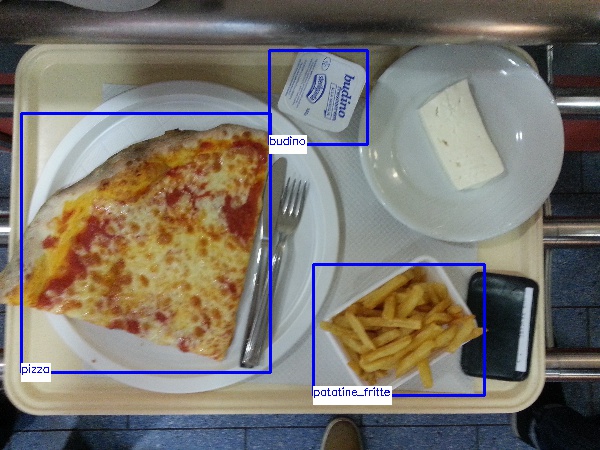} \hfill
\includegraphics[width=0.325\textwidth]{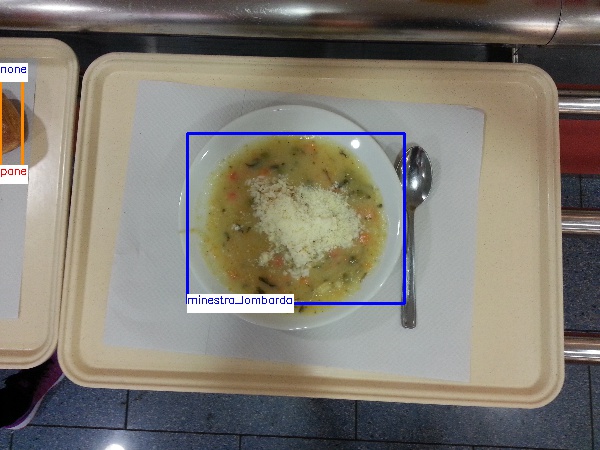} \hfill
\includegraphics[width=0.325\textwidth]{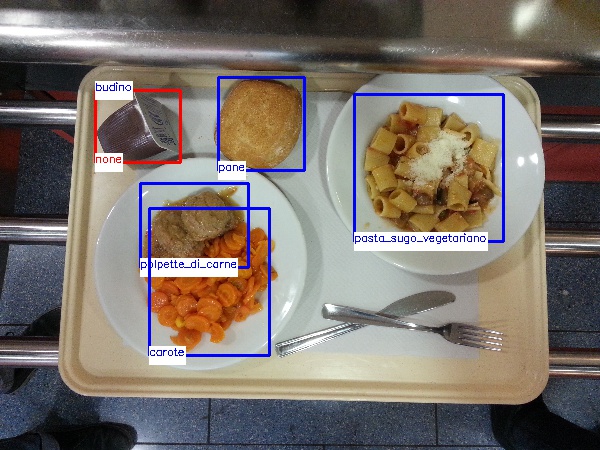} \\ \vspace*{0.2cm}
\includegraphics[width=0.325\textwidth]{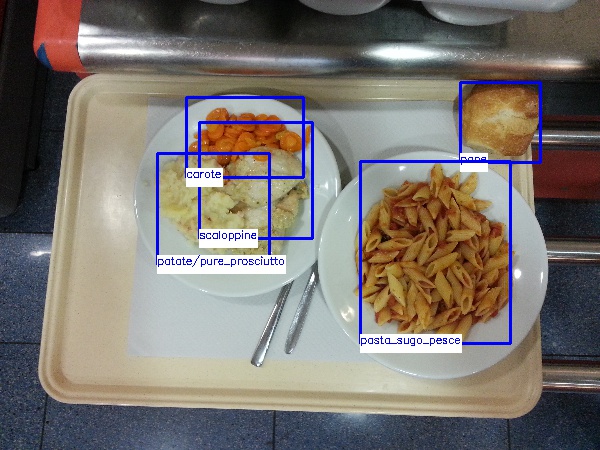} \hfill
\includegraphics[width=0.325\textwidth]{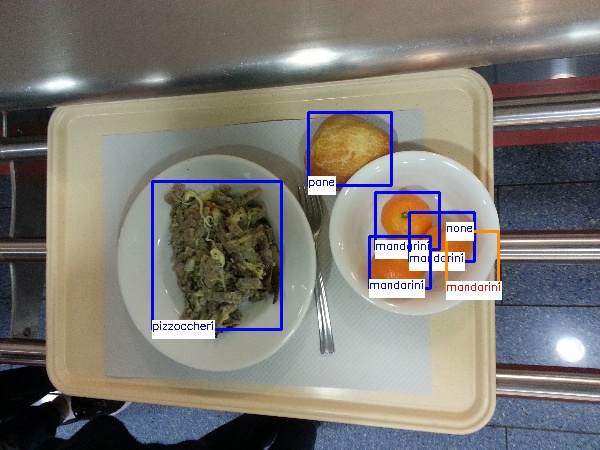} \hfill
\includegraphics[width=0.325\textwidth]{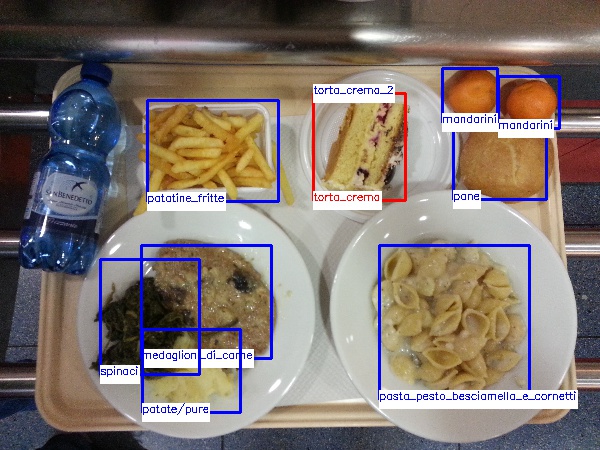} 
\caption{Some samples of the results obtained using the proposed approach, from left to right: food trays with all the objects correctly detected (blue), common false detections (orange), and misclassified objects (red).}
\label{fig:fd_results}
\end{figure*}

The results obtained with the proposed approach based on the number of objects to be classified per food tray is shown in Fig. \ref{fig:fbt_results} (b). As expected, the \textit{TA} measure tends to decrease as the number of objects increases, however there is no clear trend for the \textit{Recall}. One of the lowest results in both measure is obtained in trays containing 6 foods, whereby we can determine that the errors correspond to 17 misclassified objects along 12 trays, that is, an average error of 1.42 objects per incorrectly classified tray. Despite having a low \textit{TA} (0.478), the results are good considering the \textit{Recall} obtained, since it is preferable to minimize the number of errors per tray if we think of a semi-automatic food billing system, in which the operator would make minor corrections if necessary.


When reviewing the overall mean of errors by misclassified trays, we can see that our classifier has an average of 1.09 errors along 85 trays classified incorrectly, compared to \cite{ciocca2017food} that has an average of 3.33 errors along 76 trays classified incorrectly. That said, even though the baseline method achieves to completely classify 9 trays more than our proposal, due to its overall performance, the misclassified trays have about three times as many objects wrongly classified per tray.

Finally, Figure \ref{fig:fd_results} shows some examples of the results obtained by means of our proposed Semantic Food Detection method. In general terms, the classifier achieves a good performance in a variety of food items, where the main difficulties encountered are due to the following issues: 1) unlabeled food items, because they are not part of the 65 classes (eg. fresh cheese) or because they are not belonging to the same tray and that have been recognized by our algorithm (eg. pane), 2) the same food items placed very close (eg. mandarine), 3) foods ignored because they are not clearly distinguishable whether correspond to a meal or not (eg. pudding), and 4) confusions with classes corresponding to different kinds of cakes (eg. torta\_cream), meats, pastas, among others.



\section{Conclusion} \label{sec:conclusion}

In this paper, we presented a novel system that performs Semantic Food Detection, which combines semantic segmentation, localization and recognition techniques. We applied this methodology to the problem of food tray analysis in self-service restaurants. More precisely, we integrated several techniques: 1) food/non-food semantic segmentation through FCNs, 2) food detection, which includes localization and recognition, and 3) non-maximum suppression to avoid the occurrence of false detections. As for the results, our proposal significantly outperforms the state-of-the-art in terms of recall and mean average accuracy. 

Another aspect to emphasize is that our model is less sensitive to class imbalance, and also the mean of errors per foods placed on a tray is about 1, when the classifier does not achieve to recognize the whole tray well. The latter is quite relevant if the Semantic Food Detection is applied in a semi-automatic billing system, in which the cashier would have to make only small changes to generate the final bill, and in this way to streamline the process involved in a self-service restaurant of {\em grab a meal, pay, and eat}. Furthermore, our semantic food detection approach takes less than 0.5 seconds to predict all foods present in a image, considering the use of a personal computer with a low performance GPU (GeForce 940MX). 

Our future research is focused on semantic detection of food ingredients and completely automating the self-service billing by integrating the restaurant menu by geolocalization.

\section*{Acknowledgment}

The authors gratefully acknowledge the support of NVIDIA Corporation with the donation of the Titan Xp GPU.

\ifCLASSOPTIONcaptionsoff
  \newpage
\fi


\bibliographystyle{IEEEtran}
\bibliography{biblio}

%








\end{document}